\documentclass{article}

\usepackage{arxiv}

\usepackage[utf8]{inputenc} 
\usepackage[T1]{fontenc}    
\usepackage{hyperref}       
\usepackage{url}            
\usepackage{booktabs}       
\usepackage{amsfonts}       
\usepackage{nicefrac}       
\usepackage{microtype}      
\usepackage{graphicx}
\usepackage{natbib}
\usepackage{doi}
\usepackage{longtable}      
\usepackage{array}          
\usepackage{pdflscape}      
\usepackage{ragged2e}       
\usepackage{xurl}

\newcolumntype{L}[1]{>{\RaggedRight\arraybackslash}p{#1}}

\title{Large Language Models and Forensic Linguistics: Navigating Opportunities and Threats in the Age of Generative AI}

\date{} 					

\author{ \href{https://orcid.org/0000-0002-4093-5973}{\includegraphics[scale=0.06]{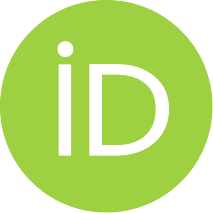}\hspace{1mm}George Mikros}\\
	College of Humanities and Social Sciences\\
    Hamad Bin Khalifa University\\
	Doha, Qatar \\
	\texttt{gmikros@hbku.edu.qa} \\
}

\renewcommand{\shorttitle}{LLMs and Forensic Linguistics}

\hypersetup{
    pdftitle={Large Language Models and Forensic Linguistics},
    pdfsubject={Forensic Linguistics, AI},
    pdfauthor={George Mikros},
    pdfkeywords={forensic linguistics, large language models, authorship attribution, AI detection},
    colorlinks=true,       
    linkcolor=blue,        
    citecolor=blue,        
    filecolor=magenta,     
    urlcolor=blue          
}

\begin{document}
\maketitle

\begin{abstract}
Large language models (LLMs) present a dual challenge for forensic linguistics. They serve as powerful analytical tools enabling scalable corpus analysis and embedding-based authorship attribution, while simultaneously destabilising foundational assumptions about idiolect through style mimicry, authorship obfuscation, and the proliferation of synthetic texts. Recent stylometric research indicates that LLMs can approximate surface stylistic features yet exhibit detectable differences from human writers, a tension with significant forensic implications. However, current AI-text detection techniques, whether classifier-based, stylometric, or watermarking approaches, face substantial limitations: high false positive rates for non-native English writers and vulnerability to adversarial strategies such as homoglyph substitution. These uncertainties raise concerns under legal admissibility standards, particularly the Daubert and Kumho Tire frameworks. The article concludes that forensic linguistics requires methodological reconfiguration to remain scientifically credible and legally admissible. Proposed adaptations include hybrid human–AI workflows, explainable detection paradigms beyond binary classification, and validation regimes measuring error and bias across diverse populations. The discipline's core insight, i.e., that language reveals information about its producer, remains valid but must accommodate increasingly complex chains of human and machine authorship.
\end{abstract}


\section{Introduction}
Forensic linguistics, broadly defined as the application of linguistic knowledge and methods to legal problems, has developed into a recognised sub-discipline since the late twentieth century \citep{coulthard2007introduction, mcmenamin2002forensic}. The field conventionally encompasses three overlapping domains: the language of legal texts, the language of legal processes, and the work of the linguist as an expert witness in cases involving authorship attribution, threatening communications, trademarks, and related phenomena \citep{coulthard2010forensic}.

Central to much forensic practice is the concept of idiolect, the idea that each individual has a relatively stable and distinctive linguistic “fingerprint” manifested across their spoken and written productions \citep{coulthard2004author}. This assumption has supported numerous cases in which linguistic evidence has been adduced to associate, or dissociate, a suspect with a disputed text.

The rapid development of large language models (LLMs) represents the most serious challenge yet to this framework. Transformer-based systems trained on massive text corpora now generate fluent and contextually appropriate language that can emulate a wide range of registers and styles \citep{kumarage2023neural}. The public deployment of ChatGPT in November 2022 and the swift adoption of similar systems have dramatically increased the volume of synthetic text in everyday communication. Within two months, ChatGPT alone attracted over 100 million monthly active users, altering the ecology of digital text production on a global scale \citep{liang2023gpt}.

In this context, earlier calls for rigorous validation and adaptation of authorship analysis as a forensic science \citep{ainsworth2019who} have acquired new urgency. The implications are not merely theoretical. Accurate authorship attribution and robust assessment of linguistic evidence remain crucial for criminal investigations, civil litigation, academic integrity proceedings, and the mitigation of misinformation \citep{huang2024authorship}. Yet the same technologies that can assist forensic analysis can also be deployed to evade it. Writers can now outsource parts of their writing, mask their stylistic idiolect, or imitate the style of others with relative ease. As a result, it is increasingly difficult to assume that any given document is the output of a single human writer operating without machine assistance.

This article, therefore, addresses the following three interrelated research questions:
\begin{enumerate}
    \item In what ways can LLMs enhance forensic linguistic methodologies and extend the analytical capacities available to practitioners?
    \item What threats do LLMs pose to traditional forensic practices, particularly to idiolect-based authorship attribution and the evaluation of textual evidence?
    \item How should the field adapt methodologically, institutionally and legally to maintain reliability and admissibility under standards such as Daubert in an era of generative AI?
\end{enumerate}

To answer these questions, this article offers a critical narrative review of research on LLMs and forensic linguistics published primarily between 2023 and 2025, supplemented by foundational work in authorship analysis and legal admissibility. The review prioritises studies directly relevant to forensic provenance tasks, including human authorship attribution, AI-text detection, model-source attribution, and human–LLM hybrid texts. Evidence was identified through targeted searches across computational linguistics, digital forensics, and law-oriented venues, with emphasis on work reporting empirical evaluations, error rates, bias, or adversarial robustness. Given the pace of model change and the prevalence of recent preprints, the findings are treated as indicators of current trajectories rather than settled consensus. The synthesis is organised through a dual-use lens, mapping how LLMs both extend forensic capacity and destabilise idiolect-based assumptions, and the legal discussion is anchored in Daubert/Kumho to translate technical limitations into evidentiary implications.

The article proceeds as follows. We begin by outlining the theoretical and methodological foundations of forensic linguistics, alongside the technical characteristics of LLMs relevant to forensic applications. Subsequent sections examine the opportunities that LLMs create as analytical tools, the threats they pose to existing practices, and current detection and countermeasure strategies. A discussion section synthesises these strands, presenting implications for practitioners and legal systems, limitations of the present analysis, and priorities for future research, before offering concluding remarks.

\section{Theoretical Framework}

\subsection{Forensic linguistics: Core principles and methods}
Authorship attribution and related forensic linguistic practices rely on two principal assumptions. The first is that each individual exhibits a distinctive idiolect. The second is that features characteristic of that idiolect recur with sufficient stability across a person's texts to permit meaningful comparison \citep{coulthard2004author, grant2007quantifying}. These assumptions follow closely those of other pattern comparison forensic sciences, in which properties of a suspect are inferred from regularities in their products \citep{ainsworth2019who}.

The theoretical foundations of idiolect have been substantially advanced by \citet{nini2023theory} Theory of Linguistic Individuality, which provides a formal framework grounded in cognitive linguistics and usage-based language processing. Nini argues that previous computational approaches to authorship analysis have implicitly adopted an untenable model of language production, one in which writers select words one by one according to grammatical rules. Drawing on \citet{christiansen2016now} Now-or-Never Bottleneck theory and \citet{langacker1987foundations} Cognitive Grammar, Nini proposes instead that language is processed in chunks: automatically produced units stored in long-term memory that range from character clusters and morphemes to multi-word sequences and schematic constructions. These units exist on a continuum of entrenchment, with more frequently used structures becoming more deeply automatised and readily accessible. A person's grammar at any point in time is thus the set of units whose entrenchment exceeds a threshold permitting automatic production, and an individual's idiolect is formally defined as the indexed family of these grammars across their lifetime, capturing the dynamic, mutable nature of linguistic knowledge \citep{nini2023theory}.

This theoretical apparatus explains why the combinatorics of language lead to individuality. As \citet{coulthard2004author} proposed, uniqueness arises not from single distinctive features but from idiolectal co-selection: the conjunction of many choices that, individually, may be unremarkable but that, in combination, form patterns potentially unique to an individual. The Unabomber case exemplifies this principle: twelve linguistic items, including phrases such as \textit{at any rate}, \textit{clearly}, and \textit{presumably}, uniquely connected the manifesto to writings by Ted Kaczynski \citep{coulthard2017introduction}. Nini's framework provides the cognitive-linguistic explanation for this phenomenon: individuals differ in their repertoires of entrenched chunks because each person's linguistic experience is unique, leading to idiosyncratic patterns of automatisation that manifest as distinctive stylistic profiles.

Historically, these assumptions have given rise to two broad methodological traditions. One is qualitative forensic stylistics, associated with close reading of texts for distinctive lexical, syntactic, and discourse-pragmatic features \citep{mcmenamin2002forensic}. The other is stylometry, which emerged from literary studies and computational linguistics and focuses on statistical measures over features such as function-word frequencies and character n-grams. The work of \citet{burrows2002delta} is particularly influential in the latter tradition. His Delta measure calculates the mean absolute difference between z-scores of frequent word types across texts, allowing for the simultaneous comparison of a disputed text with multiple candidate authors. This approach is based on the insight that unconscious, highly frequent linguistic choices—and particularly function words—are more resistant to deliberate manipulation and therefore serve as robust markers of authorial style.

\citet{evert2017understanding} demonstrated that the success of Delta-like measures depends primarily on vector normalisation and proposed the Key Profiles Hypothesis: authors are distinguished by the overall pattern of variation across features in relation to a norm, not by the magnitude of individual deviations. This insight aligns with \citet{nini2023theory} theory, suggesting that it is the configuration of an individual's entrenched units—their position in a multidimensional space of linguistic choices—that constitutes their stylometric signature. Subsequent refinements of Delta, including alternative normalisation schemes and distance metrics, have produced consistent improvements in attribution performance \citep{jannidis2015improving}.

Crucially, the accessibility of these methods to the broader research community has been greatly enhanced by dedicated software tools. The R package Stylo, developed by \citet{eder2016stylometry}, has become a standard resource for computational text analysis, providing implementations of Delta and its variants alongside functions for cluster analysis, bootstrap consensus trees, and rolling stylometry for detecting style shifts within texts. \citet{eder2013does} complementary work on sample size requirements has offered practical guidance on the minimum text lengths needed for reliable attribution, while his investigations into feature selection \citep{eder2011style} have clarified which portions of the frequency spectrum carry the most discriminatory power.

At the same time, contemporary forensic stylistics extends far beyond simple word-frequency analysis. Practitioners now routinely consider features at multiple levels of linguistic organisation, including phraseology, punctuation practices, orthographic conventions, measures of lexical diversity and syntactic complexity \citep{berriche2024unveiling}. \citet{mikros2013authorship} formalised this intuition with their Author's Multilevel N-gram Profiles (AMNP) representation, which combines character n-grams of varying lengths into a unified author model. Applied initially to Greek Twitter data, AMNP demonstrated that capturing orthographic and sub-lexical patterns at multiple granularities yields robust attribution even in the challenging conditions of social-media text, where samples are short and register is informal. Tools such as StyloMetrix \citep{okulska2023stylometrix} continue this multi-dimensional approach by extracting stylometric vectors across grammatical, lexical and syntactic dimensions that can be used in both research and casework.

In any forensic application, however, methodological sophistication is not sufficient in itself. In the United States, the admissibility of expert evidence, including linguistic evidence, is governed in federal courts by the Daubert standard set out in \textit{Daubert v. Merrell Dow Pharmaceuticals} (1993) and extended by \textit{Kumho Tire Co. v. Carmichael} (1999). Under this framework, trial judges assess whether expert testimony rests on a sufficiently reliable scientific foundation. Relevant factors include the testability of the technique, its exposure to peer review and publication, its known or potential error rate, the existence of standards controlling its application and its degree of acceptance in the relevant scientific community \citep{national2023law}. For forensic linguistics, these criteria imply that authorship attribution and related methods must be empirically validated, with clearly characterised error rates and transparent reporting, if they are to remain admissible and persuasive in court.

\subsection{Large language models: Technical foundations and forensic relevance}
Large language models represent the current state of the art in natural language processing. They are built on the transformer architecture introduced by \citet{vaswani2017attention}, which uses self-attention mechanisms to model long-range dependencies in text. These models are trained on extremely large corpora using self-supervised objectives, usually next-token prediction, and learn internal representations that encode syntactic structure, semantic relations and discourse patterns. Contemporary systems such as GPTs, Claude, LLaMA and Gemini contain billions or even trillions of parameters and exhibit emergent capabilities including in-context learning, instruction following and flexible style control \citep{kumarage2023neural}.

Several properties of LLMs are especially important for forensic linguistics. First, they are highly flexible in style. Through prompting or fine-tuning, they can generate text in particular registers, genres, or voices. Research on one-shot style transfer demonstrates that even limited authorial samples can be used to guide LLM outputs towards specific stylistic profiles \citep{miralles2025llm}. Second, in their default mode, LLMs tend towards statistically “average” patterns observed in the training data. As \citet{przystalski2025stylometry} show, LLM-generated texts typically exhibit greater grammatical standardisation than human texts. This standardisation creates detectable differences that can be exploited for classification, but it also means that LLM output often lacks the idiosyncratic irregularities on which human authorship attribution relies.

Stylometric studies suggest that, at present, LLMs have not entirely converged with human writers. \citet{osullivan2025stylometric} finds that human-authored creative texts and LLM-generated imitations occupy distinct regions of stylometric space when analysed using Burrows’ Delta. \citet{mikros2025beyond}, in a hierarchical cluster analysis of GPT-4o’s capacity for literary style imitation, reports that while the model can group its imitations of a given author together, these imitations remain separable from genuine texts by that author. \citet{wang2025catch} show that LLMs still struggle to reproduce the implicit, fine-grained writing styles of ordinary authors from limited prompts, suggesting that deeper idiosyncratic patterns remain relatively resistant to mimicry. At the same time, there is evidence that the gap is narrowing. \citet{zaitsu2025stylometry} report that outputs from more recent systems, such as GPT-o1, are stylometrically closer to human writing than outputs from earlier models such as GPT-4o in a Japanese corpus.

For forensic linguistics, the consequence is a dynamic and unstable landscape. LLMs provide new tools for analysing text at scale and for representing stylistic patterns in high-dimensional embedding spaces. However, at the same time, they threaten existing assumptions about what constitutes an individual’s “own” writing and challenge the robustness of methods that were developed for exclusively human text.

\section{Opportunities: LLMs as tools for forensic linguistics}
The first research question concerns the ways in which LLMs can function as tools that extend, rather than merely undermine, forensic linguistic practice. Recent work suggests three main areas of opportunity: enhanced authorship attribution, scalable multitask and multilingual analysis, and new forms of fine-grained and explainable stylistic analysis.

\subsection{Strengthening authorship attribution}
One line of research treats LLMs themselves as attribution engines. \citet{hu2024bayesian} propose a Bayesian framework that uses LLMs to estimate the probability that a questioned text entails the writings of a candidate author. Rather than relying only on surface features, their method frames authorship attribution as a problem of probabilistic inference over textual entailment. Using pretrained models such as Llama3-70B in one-shot settings with ten authors, they report accuracy of approximately 85\% on IMDb and blog corpora. This work illustrates how LLM-based contextual embeddings and entailment judgements can operationalise a more explicitly probabilistic conception of authorship, one that aligns well with the evidential reasoning required in forensic contexts.

\citet{choi2025find} extend the LLM-as-attributor paradigm to source code, a domain with clear analogies to forensic text analysis. They show that LLMs can assess, in a zero-shot setting, whether two code snippets come from the same author, achieving a Matthews correlation coefficient of 0.78. To handle large sets of candidate authors, they introduce a tournament-style strategy that compares subsets of authors within input length constraints, attaining classification accuracy of roughly two-thirds across hundreds of authors and multiple programming languages using only a single reference per author. Although their empirical focus is code, the general methodology of pairwise LLM-based comparison is highly relevant to forensic text attribution, where cases often involve multiple possible authors and limited reference material.

A related but distinct development concerns attribution of LLM-generated text to particular models. \citet{bisztray2025know} present CodeT5Authorship, a model trained on LLM-generated code that achieves more than 97\% accuracy in distinguishing between closely related systems such as GPT-4.1 and GPT-4o. Their LLMAuthorBench dataset comprises 32,000 AI-written computer programs and provides a benchmark for model-source attribution. While this work again focuses on code, it demonstrates that synthetic texts can carry model-specific signatures that are detectable at scale. Forensic linguistics, which increasingly must address questions about whether a given text was produced by a particular LLM, stands to benefit directly from such techniques.

Together, these studies indicate that LLMs can be used to strengthen authorship analysis in two complementary ways: by providing more powerful tools for human authorship attribution, and by enabling robust attribution of LLM-generated texts to specific models and model families.

\subsection{Scalable, multitask and multilingual analysis}
A second opportunity lies in the capacity of LLM-based systems to perform multiple provenance-related tasks jointly and to operate across languages. \citet{rao2025two} introduce DAMTL, a multitask learning framework that combines AI-text detection with LLM-source attribution. Evaluated across nine datasets and four backbone models, DAMTL learns shared representations that improve performance on both tasks while preserving task-specific distinctions. This design reflects the reality of many forensic problems, in which the question is not merely whether a text is “AI-generated,” but also which system produced it and how that information interacts with hypotheses about human involvement.

\citet{lacava2025authorship} address the multilingual dimension by formulating the problem of authorship attribution for machine-generated texts across 18 languages. Their results demonstrate that cross-lingual transfer is challenging, with performance degrading when moving between language families and scripts. These findings echo earlier empirical work on cross-linguistic stylometry. \citet{juola2016cross} demonstrated that certain stylometric features, including vocabulary richness measures, average word length, and platform-specific conventions, correlate strongly across languages for the same author, suggesting that language-independent authorship signals do exist. However, their subsequent comparative study \citep{juola2019comparative} revealed that attribution accuracy varies substantially between languages even under tightly controlled conditions: using the same authors writing on the same topics, performance on English texts was significantly higher than on Greek, a disparity attributed to differences in morphological complexity and word-order flexibility.

Nonetheless, this body of work establishes that LLM-based approaches can be applied beyond English, provided that models are carefully evaluated in each specific linguistic context. For forensic practitioners dealing with multilingual evidence, the possibility of attribution without extensive language-specific feature engineering is particularly attractive, but it also underscores the need for cautious interpretation and locally validated models.

At a broader level, LLMs support a dramatic increase in the scale at which forensic linguistic analyses can be conducted. \citet{dunsin2024comprehensive}, in a survey of AI and machine learning in digital forensics, note that these technologies allow investigators to process vast quantities of digital artefacts while maintaining acceptable levels of precision. For forensic linguistics, this implies the ability to conduct large-scale authorship profiling, threat assessment and consistency analysis over millions of documents, reserving human expertise for the most contentious or high-stakes cases.

\subsection{New analytical capabilities and explainable workflows}
The third area of opportunity concerns forms of analysis that were previously difficult or impossible. \citet{romisch2025better} examine whether LLMs can detect sentence-level style change, a task of relevance to questions of interpolation, ghostwriting and document tampering. Their results indicate that state-of-the-art LLMs can, even without task-specific fine-tuning, identify subtle stylistic shifts within a document more accurately than traditional baselines. In forensic settings, such capabilities could support the systematic examination of long texts for points at which authorship may have changed.

\citet{abbas2025attribution} compares two approaches to authorship attribution in a Human–AI parallel corpus spanning six domains: fixed style embeddings and an instruction-tuned LLM judge (GPT-4o). He reports that the LLM judge performs better for fiction and academic prose, where semantic and discourse-level features are crucial, whereas fixed embeddings perform better for dialogue-like data with more structural regularities. These findings suggest that no single method is optimal across all domains. Instead, hybrid workflows that combine multiple analytical components, each aligned with particular text types, are likely to yield the most reliable forensic conclusions.

Explainability is a further critical consideration. \citet{roemling2024explainability} explore the use of explainable machine-learning approaches in a case study on geolinguistic authorship profiling. By employing techniques such as SHAP to identify features that drive model decisions, they show that high-performing models can be made interpretable in ways that are meaningful for legal audiences. For forensic linguistics, this is particularly important: any analytical gain achieved by LLM-based methods will be of limited value if experts cannot explain, in linguistic and legally accessible terms, why a system reached a particular conclusion.

The above-mentioned studies demonstrate that LLMs can both augment established methods and open new avenues for analysis. They provide richer representations of stylistic similarity, support joint detection and attribution across languages and tasks, and enable fine-grained and explainable analyses that are better aligned with the evidentiary demands of courts.

\section{Threats: How LLMs disrupt forensic linguistics}
The second research question concerns the ways in which LLMs threaten existing forensic practices. Three sets of challenges are particularly salient: the undermining of idiolect-based assumptions through style mimicry and obfuscation; the proliferation of synthetic text and the weaknesses of current detection tools; and the resulting strain on legal and methodological foundations of the discipline.

\subsection{Style mimicry and the idiolect challenge}
LLMs make it substantially easier for writers to manipulate style, either to conceal their own identity or to impersonate others. \citet{alperin2025masks} examine such “masks and mimicry” attacks on authorship verification systems. Using LLMs to paraphrase and stylistically transform texts, they show that adversaries can substantially degrade the performance of verification models, thereby weakening the evidential value of stylistic consistency or inconsistency.

\citet{huang2024authorship}, in a comprehensive review, argue that the traditional notion of idiolect as a stable, individual stylistic signature is increasingly problematic in the LLM era. They distinguish four core tasks: attribution of human-written text, detection of AI-generated text, attribution of AI-generated text to models and attribution in human–LLM coauthored settings. Hybrid texts, where AI suggestions are incorporated, edited and recontextualised by human writers, are particularly challenging, as they no longer correspond to the relatively clean categories assumed by many earlier attribution methods.

Nevertheless, stylometric studies indicate that, at least for now, LLMs have not eliminated the forensic value of style. \citet{osullivan2025stylometric} demonstrates that human and LLM-generated creative texts form distinct clusters under Delta-based analysis, even when the models attempt to imitate specific authors. \citet{wang2025catch} similarly find that LLMs struggle to reproduce the implicit writing styles of everyday authors from limited prompts, suggesting that deeper idiosyncratic patterns remain relatively resistant to mimicry. \citet{mikros2025beyond} provides further details in his analysis of GPT-4o’s literary imitations. He reports that while GPT-4o can produce internally consistent imitations of a target author’s style, these imitations remain stylometrically separable from genuine texts by that author, and generic GPT outputs are more distinct still.

These convergent findings suggest a complex picture. LLMs substantially lower the barrier to style manipulation and so undermine naive applications of idiolect. However, at the same time, the LLM-assisted texts appear to carry their own detectable signatures, which, if appropriately modelled, may themselves support forensic inferences about authorship and AI involvement.

\subsection{Synthetic text, detection failures, and bias}
A second set of threats arises from the difficulty of reliably identifying AI-generated text. Detection errors are not distributed randomly, and they carry serious implications for fairness and due process. \citet{dalalah2023false} analyse the performance of AI detection tools in academic contexts and find that literature review sections are particularly prone to false positives, presumably because their formal and formulaic style resembles LLM output. \citet{rashidi2023chatgpt} document similar problems in medical informatics, where human-written manuscripts are sometimes misclassified as AI-generated by detection software, with false positive rates exceeding up to eight per cent for some journal types.

More troublingly, \citet{liang2023gpt} demonstrate that several widely used detectors are strongly biased against non-native English writers. In their study of seven detectors, a large proportion of TOEFL essays written by Chinese students were incorrectly labelled as AI-generated, with one detector flagging nearly 98 per cent of such essays as synthetic, while texts by native English speakers were rarely misclassified. The authors attribute this effect to the use of perplexity-based heuristics: texts with simpler, more predictable vocabulary tend to be classified as AI-generated, whereas more lexically complex texts are classified as human-authored. This design choice systematically penalises writers whose linguistic resources are constrained, thereby embedding linguistic proficiency bias into detection outcomes.

Detection systems are also vulnerable to adversarial manipulation. \citet{creo2024silverspeak} show that homoglyph substitution, i.e., replacing characters with visually similar characters from other scripts, can dramatically degrade detector performance. Across seven detection systems, they report an average drop in Matthews correlation coefficient from 0.64 to –0.01, effectively reducing classification performance to below chance while preserving human readability.

\citet{zeng2024detecting} consider sentence-level detection in human–AI collaborative texts from the CoAuthor corpus and find that detectors perform poorly when segments are short, authorship alternates frequently or AI-generated sentences have been edited by humans. Under such conditions, accurately quantifying the contribution of AI to a document becomes extremely difficult. These findings indicate that current detection tools are both biased and fragile. They misclassify human writing in ways that disproportionately affect non-native speakers and users of particular genres, and they are easily evaded by relatively simple manipulations. As such, they cannot yet serve as reliable forensic instruments in high-stakes settings.

\subsection{Legal and methodological strain}
The technical vulnerabilities described above translate directly into legal and methodological challenges. Under Daubert and \textit{Kumho Tire}, expert methods must be demonstrably reliable, with known error rates that withstand adversarial scrutiny \citep{daubert1993daubert, kumho1999kumho, national2023law}. \citet{ainsworth2019who} argue that authorship analysis can only function as a model of forensic science if it embraces rigorous validation. Given the error profiles reported by \citet{liang2023gpt} and others, it is difficult to argue that current AI-text detectors meet these requirements, particularly in contexts where decisions affect liberty, livelihood, or reputation.

In addition, the mere availability of sophisticated text generators alters the strategic landscape of litigation. A defence counsel can point to the possibility that a questioned document was produced or modified by an LLM, thereby introducing doubt about authorship or intent. Prosecutors, in turn, may find it harder to claim that stylistic alignment between an individual’s writing and a disputed text strongly supports authorship when AI-assisted alternatives are plausible. \citet{giray2024problem} shows, in the academic domain, how false accusations of AI-assisted writing disproportionately affect scholars writing in a second language or with distinctive stylistic profiles, raising concerns about discrimination and due process that are directly relevant to forensic practice.

On the methodological side, the standardisation effect of LLM output documented by \citet{przystalski2025stylometry} and others implies that measures such as Burrows’ Delta, which depend on individual variation in word-frequency distributions, may be less discriminating when applied to LLM-assisted texts. Prompting and fine-tuning add further complications. \citet{liang2023gpt} show that simple prompt modifications can substantially reduce detectability of AI-generated text. \citet{miralles2025llm} demonstrate that fine-tuning LLMs on small author corpora can produce convincing imitations of individual styles, further eroding the intuitive link between stylistic similarity and common authorship.

\citet{zhang2025multihierarchical} investigate multi-hierarchical feature detection for LLM-generated text and find that integrating additional handcrafted features yields only marginal improvements in detection performance at considerable computational cost. These results suggest that simply adding more features or complexity to existing frameworks will not, on its own, resolve the underlying challenges.

The combined effect is that both the legal and methodological bases of traditional forensic linguistics are under strain. Longstanding assumptions about idiolect, authorship and the provenance of texts can no longer be taken for granted in an environment where AI assistance is pervasive and difficult to observe.

\section{Detection and countermeasures}
The third research question concerns how forensic linguistics might adapt to the presence of LLMs. One component of such adaptation involves improving AI-text detection and developing complementary provenance mechanisms. Recent work clusters into three strands: enhancements to classifier-based and stylometric detection; watermarking and provenance infrastructures; and a reconceptualisation of detection as a fine-grained, often semi-supervised, task.

\subsection{Current detection methodologies}
Several recent studies focus on improving classifier-based detection. \citet{macko2025robustly} reports that robust fine-tuning of an LLM for binary and multiclass detection yields high accuracy across diverse datasets. \citet{trivedi2025sarang}, in the DEFACTIFY 4.0 shared task, obtain F1 scores of 1.0 for binary detection and 0.9531 for multiclass model attribution using ensembles of DeBERTa models trained with noise-augmented data. Their approach illustrates the value of modelling distributional variation and corruption that may arise in real-world settings.

Stylometric detection offers a complementary perspective. \citet{przystalski2025stylometry} use features from StyloMetrix and n-gram statistics to distinguish human from LLM texts across several model families, including GPT-3.5/4, LLaMA 2/3, Orca and Falcon. They report Matthews correlation coefficients up to 0.87 in seven-class tasks and accuracies approaching 0.98 in balanced binary comparisons of human texts and GPT-4 outputs. Importantly, they employ SHAP to identify features that contribute most to classification, such as overrepresentation of particular words, punctuation patterns and markers of grammatical standardisation, thus linking model behaviour back to linguistically interpretable phenomena.

\citet{mikros2023ai} further advance this hybrid approach by demonstrating the efficacy of combining transformer-based models with extensive stylometric feature engineering. In their participation in the AuTexTification shared task, they employed an ensemble method utilizing majority voting that fused predictions from ELECTRA and RoBERTa transformers with a comprehensive set of stylometric features, including author multilevel n-gram profiles (AMNP) and LIWC indices. Their ensemble model achieved a macro-F1 score of 60.78 on the English subtask for binary classification and 55.87 for multiclass model attribution, positioning them above the median of competing teams. Their results underscore that while transformers capture subtle data patterns, stylometric features, such as punctuation, phraseology, and lexical diversity, offer complementary, linguistically grounded signals that enhance detection accuracy, particularly when fused in an ensemble framework.

\citet{berriche2024unveiling} adopt a different stylometric strategy, using intrinsic features with classical classifiers and ensembles such as k-nearest neighbours, decision trees, Naïve Bayes and XGBoost. Despite their relative simplicity, these models achieve competitive performance on ChatGPT detection when features are carefully engineered and tuned. \citet{zaitsu2025stylometry}, in a Japanese context, combine phrase patterns, part-of-speech bigrams, and function-word unigrams to discriminate between human texts and texts generated by seven LLMs. They find that stylometric features can, in some settings, perfectly separate machine from human texts and reveal clustering among models themselves, with Llama 3.1 exhibiting particularly distinctive characteristics.

These studies show that, under controlled conditions, both classifier-based and stylometric approaches can perform impressively. However, as discussed previously, their performance deteriorates in adversarial, short-text, or hybrid contexts, and their fairness properties remain a matter of concern.

\subsection{Watermarking and provenance techniques}
Watermarking seeks to embed imperceptible patterns in LLM output that can later be used to verify that a text was machine-generated. \citet{kirchenbauer2023reliability} describe a scheme in which token sampling is biased to favour tokens from a “green list” while preserving fluency. A detector with access to the secret key can then test whether a text exhibits the expected statistical signature. This approach has the advantage of not requiring changes to model architecture or training and, in principle, could be deployed broadly.

\citet{verma2025watermarking}, however, show that watermarking can degrade model alignment, reducing properties such as truthfulness, safety, and helpfulness. They argue that there is a fundamental tradeoff between watermark strength and output quality. \citet{xu2025mark} focus on watermarking as a means to detect misuse of open-source LLMs, such as unauthorised inclusion in downstream systems. \citet{cao2025watermarking} reviews watermarking methods across text, image, and audio and emphasises that, even when technically sound, watermarking requires adoption by model providers and is vulnerable to removal or distortion via paraphrasing and other post-processing operations. \citet{zhang2025cohemark} propose CoheMark, a sentence-level watermark designed to preserve or improve perceived text quality while remaining detectable under common edits. Their results suggest that careful tuning can ameliorate some of the alignment–quality tradeoffs identified by \citet{verma2025watermarking}.

Beyond watermarking, content provenance initiatives such as the Coalition for Content Provenance and Authenticity (C2PA) seek to establish cryptographic chains of custody for digital artefacts, potentially including text. In combination, watermarking and provenance systems could provide robust evidence about the involvement of particular models or platforms in document creation, though such infrastructures are still in early stages of deployment.

\subsection{Finegrained and semisupervised detection}
A recurrent theme in the literature is that binary “AI versus human” classification is rarely adequate for forensic purposes. \citet{cheng2024beyond} therefore propose a fine-grained detection paradigm that introduces two new tasks. The first, LLM Role Recognition (LLMRR), seeks to identify the role that an LLM played in content generation (for example, primary author, editor or paraphraser). The second, LLM Influence Measurement (LLMIM), aims to estimate the extent of LLM involvement on a continuous scale. Using the LLMDetect benchmark, including the Hybrid News Detection Corpus, they show that fine-tuned pretrained language models outperform other baselines on both tasks. At the same time, they note that even advanced LLMs struggle to detect their own contributions accurately. This framework better reflects the questions that often arise in legal contexts, where the degree of AI involvement may be as important as its presence.

\citet{qazi2024gpt} address a complementary problem: the difficulty of maintaining supervised detectors as new models appear. Their GpTen method is semi-supervised and requires only human-generated text for training. Evaluated on the GPT Reddit Dataset (GRiD), which contains prompts and responses from both humans and ChatGPT, GpTen performs comparably to fully supervised baselines. Semi-supervised and unsupervised approaches of this kind may be crucial in contexts where it is impractical to obtain labelled examples of output from every new model and configuration.

Collectively, these developments move detection research toward tasks and representations that are more compatible with forensic needs, emphasising roles, degrees of involvement and robustness to distributional change rather than binary classification alone.

Table 1 consolidates the paper’s core claims across Sections 3–5 by aligning real forensic questions with the dual-use affordances of LLMs, the specific failure modes that threaten evidentiary reliability, and the most defensible adaptation strategies. The table also illustrates why the field should move away from binary detection as a default forensic endpoint and toward multi-method, role-sensitive, and validation-rich approaches that are more compatible with admissibility demands.

\clearpage 
\newgeometry{margin=1cm} 
\begin{landscape}
\thispagestyle{empty} 
\pagestyle{empty}     

\footnotesize 

\begin{longtable}{L{2.5cm} L{3cm} L{3.5cm} L{3.5cm} L{3.5cm} L{3cm} L{3.5cm}}
\caption{Mapping LLM-era forensic provenance tasks to core case questions, dual-use opportunities, emerging threats, current methodological approaches, and recommended adaptations for robust and legally defensible practice.}\\
\toprule
\textbf{Forensic task} & \textbf{Typical case question} & \textbf{LLM-enabled opportunity} & \textbf{Primary LLM-era threat} & \textbf{Current approaches (illustrative)} & \textbf{Known weaknesses} & \textbf{Recommended forensic adaptation} \\
\midrule
\endfirsthead
\toprule
\textbf{Forensic task} & \textbf{Typical case question} & \textbf{LLM-enabled opportunity} & \textbf{Primary LLM-era threat} & \textbf{Current approaches (illustrative)} & \textbf{Known weaknesses} & \textbf{Recommended forensic adaptation} \\
\midrule
\endhead
\bottomrule
\endfoot
Human authorship attribution (traditional) & “Did X write this text?” & Embedding-based similarity and probabilistic/ LMM-judge paradigms can complement stylometry and qualitative analysis & Style obfuscation, paraphrase attacks, AI-assisted drafting blur author boundaries & Hybrid stylometry + transformer/ LLM-judges; Bayesian framing & Performance drops in short, noisy, or hybrid texts; domain sensitivity & Treat AI-involvement as a competing hypothesis; report uncertainty; require genre-matched validation \\
\midrule
Authorship verification & “Are these texts by the same person?” & LLMs can assist pairwise comparators and triage candidate sets & LLM-driven mimicry degrades verification baselines & Pairwise LLM assessment; stylometric verification & Vulnerable to adversarial rewriting; limited interpretability & Use multi-method convergence; explicitly test obfuscation-resistance \\
\midrule
AI-text detection (binary) & “Is this AI-generated?” & High accuracy in controlled benchmarks & False positives (esp. L2 writers); adversarial evasion (e.g., homoglyphs); hybrid texts & Classifier-based detection; stylometric feature sets & Bias and fairness issues; poor robustness to edits or short segments & Avoid sole reliance for high-stakes decisions; present error rates by demographic/ linguistic group \\
\midrule
Model-source attribution (AI-to-model) & “Which model likely produced this?” & Synthetic texts may carry model-specific signatures & Rapid model evolution may erase signatures; open-source fine-tuning complicates source claims & Multi-class detectors; stylometric/ model fingerprints & Requires up-to-date labelled data; unstable across versions & Use version-aware claims; maintain rolling benchmarks \\
\midrule
Human–LLM co-authorship analysis & “What is the degree/ role of AI involvement?” & Role-recognition and influence estimation align with real forensic questions & Role ambiguity and editing obscure traces & Fine-grained role \& involvement frameworks; segment-level analysis & Limited performance in alternating or heavily edited texts & Shift reporting from binary labels to graded involvement with confidence bounds \\
\midrule
Style-change / interpolation detection & “Was this document patched or ghostwritten?” & Sentence-level style-change detection can flag internal shifts & LLM smoothing may mask human transitions; multi-author + AI mixes complicate signals & LLM-assisted style-change detectors; rolling stylometry & False alarms in topic/ genre shifts; needs careful calibration & Combine discourse/ context evidence with stylistometric signals \\
\midrule
Watermarking (where available) & “Can we verify LLM use via embedded markers?” & Potentially strong machine-verifiable provenance & Adoption dependence; paraphrase/ translation removal; quality/ alignment trade-offs & Statistical watermark detectors & Not universal; can be stripped or avoided & Treat as supportive, not dispositive; document removal-resistance tests \\
\midrule
Cryptographic provenance / ecosystem controls & “Is there a chain-of-custody for text creation?” & Shifts detection → verification, reducing inference burden & Limited deployment; platform fragmentation & Platform-based provenance frameworks & Coverage gaps & Advocate policy/ industry standards; integrate into forensic reporting where possible \\
\end{longtable}
\end{landscape}

\restoregeometry 
\pagestyle{fancy} 

\section{Discussion}

\subsection{Synthesis of opportunities and threats}
The preceding sections have shown that LLMs simultaneously expand and constrain what forensic linguistics can do. As tools, they offer more powerful and flexible methods for authorship attribution, support joint detection and attribution across languages and tasks, and enable fine-grained, explainable analyses at scale. As sources of risk, they make style easier to manipulate, saturate communicative spaces with synthetic text, and expose the limitations of current detection tools, particularly with respect to bias and adversarial robustness.

The empirical literature on stylometry and detection suggests that LLM-generated texts still differ systematically from human writing and that these differences can often be exploited for classification and attribution under controlled conditions. At the same time, studies of detector bias, adversarial attacks, and hybrid human–AI documents demonstrate that seemingly robust methods can fail precisely in the settings that most closely resemble real forensic problems. The net result is a landscape in which opportunities and threats are tightly intertwined: the same architectures that enable improved attribution and detection also generate the texts that challenge those very systems.

\subsection{Addressing the research questions}
Viewed through the lens of the first research question, concerning the enhancement of forensic methodology, the evidence indicates that LLMs can significantly extend the analytical capacities of forensic linguistics. Bayesian and embedding-based attribution approaches, multitask frameworks for detection and model-source attribution, and sentence-level style-change detection all demonstrate that LLMs can be harnessed to perform tasks that were previously either infeasible or substantially more limited in scope \citep{hu2024bayesian, choi2025find, rao2025two, romisch2025better}. These advances sit alongside, rather than replace, traditional stylometric and qualitative methods, and they open up possibilities for hybrid human–AI workflows in which each component compensates for the other’s weaknesses.

In relation to the second research question, concerning threats to traditional practice, the studies reviewed make clear that LLMs pose a serious challenge to idiolect-based authorship attribution and to the reliability of AI-text detection. The ease of style mimicry and obfuscation, the high and uneven false-positive rates of existing detectors and their vulnerability to simple manipulations, as well as the difficulty of analysing human–AI hybrid texts, all undermine the straightforward application of existing methods \citep{alperin2025masks, liang2023gpt, creo2024silverspeak, zeng2024detecting}. At the same time, stylometric evidence that LLM-generated texts retain distinctive signatures provides some grounds for cautious optimism that new methods can be developed to cope with these challenges.

The third research question asks how the field should adapt to maintain reliability and legal admissibility in light of these developments. The detection and countermeasure literature points towards emerging strategies, including more robust classifier and stylometric approaches, watermarking and provenance mechanisms, and fine-grained, often semi-supervised detection paradigms that move beyond binary classification \citep{macko2025robustly, przystalski2025stylometry, kirchenbauer2023reliability, cheng2024beyond, qazi2024gpt}. However, the existing studies also highlight significant gaps. Many methods have not yet been validated on the kinds of short, noisy and hybrid texts that are typical of forensic casework, nor have they been systematically evaluated for fairness across linguistic and demographic groups. The subsequent sections on implications, limitations and future directions take up these findings and develop them into a more explicitly normative account of how forensic linguistics should respond.

\section{Implications for practice and law}

\subsection{Implications for practitioners}
Forensic linguists will need to extend their methodological repertoire if they are to work effectively in an AI-saturated environment. Competence in qualitative analysis and traditional stylometry remains essential, but it is no longer sufficient. Practitioners must also understand the basic operation, strengths and limitations of LLMs and detection models, including their bias profiles and susceptibility to adversarial manipulation.

Hybrid analytical strategies, in which LLM-based tools are used to generate hypotheses, identify patterns or perform initial screening, and human experts conduct contextual interpretation and final assessment, are likely to be particularly productive. Training programmes in forensic linguistics should therefore incorporate AI literacy alongside linguistics and law. This includes instruction in embedding-based and Bayesian attribution methods \citep{hu2024bayesian, abbas2025attribution}, multitask detection and attribution frameworks \citep{rao2025two}, stylometric detection techniques \citep{przystalski2025stylometry, berriche2024unveiling} and the basics of explainable AI \citep{roemling2024explainability}.

Continuing professional development will be necessary for established practitioners, given the rapid pace of technological change. Professional organisations such as the International Association for Forensic and Legal Linguistics can play a central role in articulating best practices, organising training and updating ethical guidelines to address AI-related issues. Interdisciplinary collaboration will also become increasingly important. Many of the most promising developments, such as multi-task detection and attribution, fine-grained influence estimation, and watermarking, emerge from collaborations between computer scientists, linguists, and legal scholars. Forensic linguistics programmes and laboratories would benefit from establishing institutional links with departments of computer science, information science and law to facilitate such collaborations.

\subsection{Implications for legal systems}
Legal systems must likewise adapt to the presence of LLMs. Judges and lawyers require a basic understanding of what LLMs can and cannot do, how AI-text detectors work and where they fail, and what can reasonably be inferred from stylometric or model-based analyses. Under Daubert and related standards, courts should scrutinise not only the face-value accuracy claims of proposed methods but also their validation bases, error rates across relevant populations and susceptibility to bias. Evidence that detectors systematically misclassify the writing of non-native speakers or other vulnerable groups should be of particular concern from the standpoint of due process and equal protection.

Policy interventions may include requirements for disclosure of AI assistance in certain domains, such as legal drafting, academic submissions, or regulatory filings. If such disclosures are implemented and enforced, the forensic task may shift from detection to verification, which is typically more tractable. Cryptographic provenance systems and watermarking schemes, if widely adopted, could facilitate this shift. In their absence, however, forensic linguistics will bear a significant burden in attempting to distinguish human from AI-assisted content post hoc.

The emerging fine-grained detection paradigms discussed above also have legal implications. Courts may need to move away from treating AI involvement as a simple binary condition and instead consider degrees and modes of involvement when assessing authorship, originality and responsibility. Expert testimony might then focus on reconstructing the likely contribution of human and machine actors within a collaborative writing process, rather than merely testifying that a text is or is not “AI-generated”.

\section{Limitations}
The analysis presented here is subject to several limitations. First, the empirical literature on LLMs and forensic linguistics is rapidly evolving. Many of the studies cited are recent preprints, and their findings may be superseded by subsequent work or by the release of new model generations. The observation by \citet{zaitsu2025stylometry} that newer models produce texts closer to human writing than earlier models illustrates the temporal instability of the phenomena under discussion.

Second, much of the available research focuses on English or a small number of high-resource languages, and on limited genres such as academic prose, short essays and creative writing. Forensic casework, by contrast, often involves informal, multilingual, code-switched or otherwise atypical texts. Caution is therefore needed in extrapolating from current benchmarks to real-world forensic contexts.

Third, the legal analysis in this article is framed primarily in relation to U.S. federal evidence law. Other jurisdictions apply different standards to expert evidence, and legal cultures vary in their receptivity to probabilistic and computational methods. The implications of LLMs for forensic linguistics may therefore differ across legal systems. Comparative empirical research on judicial responses to AI-related linguistic evidence remains sparse.

Finally, the article focuses on textual evidence and does not address multimodal LLMs that process speech, images, or other modalities, even though such systems may soon become relevant to forensic practice.

\section{Future directions}

\subsection{Research priorities}
Several research directions appear especially urgent. One is the development of detection methods with well-characterised error rates and fairness properties across demographic and linguistic groups. Given the documented bias against non-native English writers in current detectors \citep{liang2023gpt}, fairness-aware design and validation should be treated as a core requirement rather than a secondary consideration.

A second priority is longitudinal research tracking how LLM stylistic signatures evolve over time. As models are updated and training practices change, the features that distinguish human and machine writing today may become obsolete. Regularly updated benchmarks that include outputs from new models and diverse human authors would help maintain the relevance of detection and attribution methods.

A third priority is the systematic study of human–AI collaborative writing practices. Empirical work is needed on how authors use LLMs in different contexts, how such use affects their idiolects and what stylistic markers remain detectable in hybrid texts. The work of \citet{zeng2024detecting} on sentence-level detection in collaborative texts provides a starting point, but much remains to be understood about collaborative dynamics, editing patterns and the resulting forensic traces.

\subsection{Methodological innovation}
On the methodological side, future work is likely to focus on hybrid systems that combine multiple analytical components. LLM-based judges, embedding-based similarity measures, stylometric features and traditional qualitative analysis each contribute different strengths. Research should therefore investigate principled ways of combining them, with explicit modelling of uncertainty and error propagation. \citet{abbas2025attribution} and \citet{przystalski2025stylometry} provide initial examples of such hybrid strategies.

Explainable AI will remain central. Forensic applications require methods whose internal operations and outputs can be translated into linguistically meaningful explanations for courts. Techniques such as SHAP and other feature-attribution methods, when integrated with robust linguistic theory and empirical evidence, offer promising avenues for bridging the gap between complex models and legal discourse \citep{roemling2024explainability}.

Fine-grained detection paradigms such as those proposed by \citet{cheng2024beyond} also warrant further development and validation. Tasks that explicitly model the roles and degrees of AI involvement in text production align more closely with legal and ethical questions than simple binary detection. Research is needed to determine how well such models perform across domains, languages and collaboration modes, and how their outputs can be communicated in ways that are useful for courts and investigators.

\subsection{Policy and governance}
Finally, forensic linguistics should engage proactively with policy debates on AI governance. Questions about mandatory disclosure of AI assistance, standardisation of provenance mechanisms, regulation of watermarking and the protection of individuals from unjustified accusations of AI use are not purely technical. They implicate civil rights, academic freedom, and due process. Professional bodies in forensic linguistics, in collaboration with legal, technical, and civil-society stakeholders, can contribute expertise to the design of policies that support both the effective investigation of wrongdoing and the protection of legitimate AI-assisted writing practices.

International coordination will be particularly important. Digital texts routinely cross jurisdictional boundaries, and AI systems are developed and deployed transnationally. Shared standards for AI-related disclosures, provenance mechanisms and the evaluation of AI-mediated evidence would facilitate cross-border cooperation and reduce the risk of inconsistent or unjust outcomes.

\section{Conclusion}
Large language models present forensic linguistics with a profound set of challenges and opportunities. As analytical tools, they enable sophisticated authorship attribution methods, multitask and multilingual analysis at scale, and fine-grained stylistic detection and explanation. As generative systems, they make it easier to manipulate style, produce synthetic texts that are difficult to detect and undermine traditional assumptions about idiolect and authorship.

The evidence reviewed in this article suggests that current AI-text detection tools are not yet suitable for high-stakes forensic applications. Their high false positive rates for certain populations, their vulnerability to simple adversarial attacks and their difficulty in handling hybrid texts mean that they often fail to satisfy Daubert-style requirements for reliability. At the same time, stylometric and multitask research indicates that LLM-generated text still differs in systematic ways from human writing and that more comprehensive detection paradigms can capture aspects of AI involvement more effectively than binary classification.

The future of forensic linguistics in the LLM era will depend on the discipline’s willingness to adapt. Hybrid methodologies that combine traditional stylistic analysis, LLM-based tools, stylometric modelling and explainable AI offer one promising path. Robust validation, with explicit attention to bias and fairness, must become non-negotiable. Interdisciplinary collaboration and engagement with legal and policy frameworks will be essential to ensure that linguistic evidence remains both scientifically credible and legally admissible.

The core premise of forensic linguistics, i.e., that language use provides evidence about its producer, remains sound. What has changed is the complexity of the production process and the range of potential human and machine contributors to any given text. Meeting this complexity will require sustained efforts from researchers, practitioners, courts and policymakers. The stakes, in terms of individual rights, institutional legitimacy and the integrity of legal decision-making, are substantial. The discipline now faces a choice between resisting change and risking marginalisation, or embracing methodological and institutional reform and helping to shape a more robust and equitable forensic response to the age of generative AI.


\begin{thebibliography}{56}
\providecommand{\natexlab}[1]{#1}
\providecommand{\url}[1]{\texttt{#1}}
\expandafter\ifx\csname urlstyle\endcsname\relax
  \providecommand{\doi}[1]{doi: #1}\else
  \providecommand{\doi}{doi: \begingroup \urlstyle{rm}\Url}\fi

\bibitem[Coulthard and Johnson(2007)]{coulthard2007introduction}
Malcolm Coulthard and Alison Johnson.
\newblock \emph{An introduction to forensic linguistics: Language in evidence}.
\newblock Routledge, 2007.

\bibitem[McMenamin(2002)]{mcmenamin2002forensic}
Gerald~R McMenamin.
\newblock \emph{Forensic linguistics: Advances in forensic stylistics}.
\newblock CRC Press, 2002.

\bibitem[Coulthard(2010)]{coulthard2010forensic}
Malcolm Coulthard.
\newblock Forensic linguistics: The application of language description in legal contexts.
\newblock \emph{Langage et Soci{\'e}t{\'e}}, 132\penalty0 (2):\penalty0 15--33, 2010.
\newblock \doi{10.3917/ls.132.0015}.

\bibitem[Coulthard(2004)]{coulthard2004author}
Malcolm Coulthard.
\newblock Author identification, idiolect, and linguistic uniqueness.
\newblock \emph{Applied Linguistics}, 25\penalty0 (4):\penalty0 431--447, 2004.
\newblock \doi{10.1093/applin/25.4.431}.

\bibitem[Kumarage and Liu(2023)]{kumarage2023neural}
Tharindu Kumarage and Huan Liu.
\newblock Neural authorship attribution: Stylometric analysis on large language models.
\newblock arXiv preprint, 2023.
\newblock URL \url{https://arxiv.org/abs/2308.07305}.

\bibitem[Liang et~al.(2023)Liang, Yuksekgonul, Mao, Wu, and Zou]{liang2023gpt}
Weixin Liang, Mert Yuksekgonul, Yining Mao, Eric Wu, and James Zou.
\newblock Gpt detectors are biased against non-native english writers.
\newblock \emph{Patterns}, 4\penalty0 (7):\penalty0 100779, 2023.
\newblock \doi{10.1016/j.patter.2023.100779}.

\bibitem[Ainsworth and Juola(2019)]{ainsworth2019who}
Janet Ainsworth and Patrick Juola.
\newblock Who wrote this: Modern forensic authorship analysis as a model for valid forensic science.
\newblock \emph{Washington University Law Review}, 96\penalty0 (5):\penalty0 1161--1189, 2019.

\bibitem[Huang et~al.(2024)Huang, Chen, and Shu]{huang2024authorship}
Baixiang Huang, Canyu Chen, and Kai Shu.
\newblock Authorship attribution in the era of llms: Problems, methodologies, and challenges.
\newblock arXiv preprint, 2024.
\newblock URL \url{https://arxiv.org/abs/2408.08946}.

\bibitem[Grant(2007)]{grant2007quantifying}
Tim Grant.
\newblock Quantifying evidence in forensic authorship analysis.
\newblock \emph{International Journal of Speech, Language and the Law}, 14\penalty0 (1):\penalty0 1--25, 2007.
\newblock \doi{10.1558/ijsll.v14i1.1}.

\bibitem[Nini(2023)]{nini2023theory}
Andrea Nini.
\newblock \emph{A theory of linguistic individuality for authorship analysis}.
\newblock Cambridge University Press, 2023.
\newblock \doi{10.1017/9781108974851}.

\bibitem[Christiansen and Chater(2016)]{christiansen2016now}
Morten~H Christiansen and Nick Chater.
\newblock The now-or-never bottleneck: A fundamental constraint on language.
\newblock \emph{Behavioral and Brain Sciences}, 39:\penalty0 e62, 2016.
\newblock \doi{10.1017/S0140525X1500031X}.

\bibitem[Langacker(1987)]{langacker1987foundations}
Ronald~W Langacker.
\newblock \emph{Foundations of cognitive grammar: Vol. 1. Theoretical prerequisites}.
\newblock Stanford University Press, 1987.

\bibitem[Coulthard et~al.(2017)Coulthard, Johnson, and Wright]{coulthard2017introduction}
Malcolm Coulthard, Alison Johnson, and David Wright.
\newblock \emph{An introduction to forensic linguistics: Language in evidence}.
\newblock Routledge, 2nd edition, 2017.

\bibitem[Burrows(2002)]{burrows2002delta}
John~F Burrows.
\newblock ‘delta’: A measure of stylistic difference and a guide to likely authorship.
\newblock \emph{Literary and Linguistic Computing}, 17\penalty0 (3):\penalty0 267--287, 2002.
\newblock \doi{10.1093/llc/17.3.267}.

\bibitem[Evert et~al.(2017)Evert, Proisl, Jannidis, Reger, Pielström, Schöch, and Vitt]{evert2017understanding}
Stefan Evert, Thomas Proisl, Fotis Jannidis, Isabella Reger, Steffen Pielström, Christof Schöch, and Thorsten Vitt.
\newblock Understanding and explaining delta measures for authorship attribution.
\newblock \emph{Digital Scholarship in the Humanities}, 32\penalty0 (suppl\_2):\penalty0 ii4--ii16, 2017.
\newblock \doi{10.1093/llc/fqx023}.

\bibitem[Jannidis et~al.(2015)Jannidis, Pielström, Schöch, and Vitt]{jannidis2015improving}
Fotis Jannidis, Steffen Pielström, Christof Schöch, and Thorsten Vitt.
\newblock Improving burrows’ delta – an empirical evaluation of text distance measures.
\newblock In \emph{Proceedings of the Digital Humanities Conference 2015}. Alliance of Digital Humanities Organizations, 2015.

\bibitem[Eder et~al.(2016)Eder, Kestemont, and Rybicki]{eder2016stylometry}
Maciej Eder, Mike Kestemont, and Jan Rybicki.
\newblock Stylometry with r: A package for computational text analysis.
\newblock \emph{The R Journal}, 8\penalty0 (1):\penalty0 107--121, 2016.

\bibitem[Eder(2013)]{eder2013does}
Maciej Eder.
\newblock Does size matter? authorship attribution, small samples, big problem.
\newblock \emph{Digital Scholarship in the Humanities}, 30\penalty0 (2):\penalty0 167--182, 2013.
\newblock \doi{10.1093/llc/fqt066}.

\bibitem[Eder(2011)]{eder2011style}
Maciej Eder.
\newblock Style-markers in authorship attribution a cross-language study of the authorial fingerprint.
\newblock \emph{Studies in Polish Linguistics}, 6\penalty0 (1):\penalty0 99--114, 2011.

\bibitem[Berriche and Larabi-Marie-Sainte(2024)]{berriche2024unveiling}
Lamia Berriche and Souad Larabi-Marie-Sainte.
\newblock Unveiling chatgpt text using writing style.
\newblock \emph{Heliyon}, 10\penalty0 (12):\penalty0 e32976, 2024.
\newblock \doi{10.1016/j.heliyon.2024.e32976}.

\bibitem[Mikros and Perifanos(2013)]{mikros2013authorship}
George Mikros and K~Perifanos.
\newblock Authorship attribution in greek tweets using multilevel author’s n-gram profiles.
\newblock In \emph{Papers from the 2013 AAAI Spring Symposium "Analyzing Microtext"}, pages 17--23. AAAI Press, 2013.

\bibitem[Okulska et~al.(2023)Okulska, Stetsenko, Kołos, Karlińska, Głąbińska, and Nowakowski]{okulska2023stylometrix}
Iga Okulska, Danylo Stetsenko, Anna Kołos, Agnieszka Karlińska, Krystyna Głąbińska, and Artur Nowakowski.
\newblock Stylometrix: An opensource multilingual tool for representing stylometric vectors.
\newblock arXiv preprint, 2023.
\newblock URL \url{https://arxiv.org/abs/2309.12810}.

\bibitem[{National Institute of Justice}(n.d.)]{national2023law}
{National Institute of Justice}.
\newblock Law 101: Legal guide for the forensic expert – daubert and kumho decisions, n.d.
\newblock URL \url{https://nij.ojp.gov/nij-hosted-online-training-courses/law-101-legal-guide-forensic-expert/pretrial/pretrial-rules-evidence/daubert-and-kumho-decisions}.

\bibitem[Vaswani et~al.(2017)Vaswani, Shazeer, Parmar, Uszkoreit, Jones, Gomez, Kaiser, and Polosukhin]{vaswani2017attention}
Ashish Vaswani, Noam Shazeer, Niki Parmar, Jakob Uszkoreit, Llion Jones, Aidan~N Gomez, Łukasz Kaiser, and Illia Polosukhin.
\newblock Attention is all you need.
\newblock In \emph{Advances in Neural Information Processing Systems 30}, pages 5998--6008. Curran Associates, 2017.

\bibitem[Miralles-González et~al.(2025)Miralles-González, Huertas-Tato, Martín, and Camacho]{miralles2025llm}
Pedro Miralles-González, Javier Huertas-Tato, Alejandro Martín, and David Camacho.
\newblock Llm one-shot style transfer for authorship attribution and verification.
\newblock arXiv preprint, 2025.
\newblock URL \url{https://arxiv.org/abs/2510.13302}.

\bibitem[Przystalski et~al.(2025)Przystalski, Argasiński, Grabska-Gradzińska, and Ochab]{przystalski2025stylometry}
Karol Przystalski, Jan~K Argasiński, Inez Grabska-Gradzińska, and Jeremi~K Ochab.
\newblock Stylometry recognizes human and llm-generated texts in short samples.
\newblock \emph{Expert Systems with Applications}, 296:\penalty0 129001, 2025.
\newblock \doi{10.1016/j.eswa.2025.129001}.

\bibitem[O’Sullivan(2025)]{osullivan2025stylometric}
James O’Sullivan.
\newblock Stylometric comparisons of human versus ai-generated creative writing.
\newblock \emph{Humanities and Social Sciences Communications}, 12:\penalty0 1708, 2025.
\newblock \doi{10.1057/s41599-025-05986-3}.

\bibitem[Mikros(2025)]{mikros2025beyond}
George Mikros.
\newblock Beyond the surface: Stylometric analysis of gpt-4o’s capacity for literary style imitation.
\newblock \emph{Digital Scholarship in the Humanities}, 40\penalty0 (2):\penalty0 587--601, 2025.
\newblock \doi{10.1093/dsh/fqaf035}.

\bibitem[Wang et~al.(2025)Wang, Tripto, Park, Li, and Zhou]{wang2025catch}
Zhen Wang, Nafis~Irtiza Tripto, Seunghyun Park, Zhichao Li, and Jie Zhou.
\newblock Catch me if you can? not yet: Llms still struggle to imitate the implicit writing styles of everyday authors.
\newblock arXiv preprint, 2025.
\newblock URL \url{https://arxiv.org/abs/2509.14543}.

\bibitem[Zaitsu et~al.(2025)Zaitsu, Jin, Ishihara, Tsuge, and Inaba]{zaitsu2025stylometry}
Wataru Zaitsu, Mingzhe Jin, Shunichi Ishihara, Satoru Tsuge, and Michimasa Inaba.
\newblock Stylometry can reveal artificial intelligence authorship, but humans struggle: A comparison of human and seven large language models in japanese.
\newblock \emph{PLOS ONE}, 20\penalty0 (10):\penalty0 e0335369, 2025.
\newblock \doi{10.1371/journal.pone.0335369}.

\bibitem[Hu et~al.(2024)Hu, Zheng, and Huang]{hu2024bayesian}
Zhiqiang Hu, Tianshu Zheng, and Hong Huang.
\newblock A bayesian approach to harnessing the power of llms in authorship attribution.
\newblock arXiv preprint, 2024.
\newblock URL \url{https://arxiv.org/abs/2410.21716}.

\bibitem[Choi et~al.(2025)Choi, Tan, Meng, Ragab, Mondal, Mohaisen, and Aung]{choi2025find}
Seungwon Choi, Yan~Kiat Tan, Michael~Hwa Meng, Mohamed Ragab, Sudipta Mondal, David Mohaisen, and Khin Mi~Mi Aung.
\newblock I can find you in seconds! leveraging large language models for code authorship attribution.
\newblock arXiv preprint, 2025.
\newblock URL \url{https://arxiv.org/abs/2501.08165}.

\bibitem[Bisztray et~al.(2025)Bisztray, Cherif, Dubniczky, Gruschka, Borsos, Ferrag, Kovacs, Mavroeidis, and Tihanyi]{bisztray2025know}
Tamas Bisztray, Bechara Cherif, Richard~A Dubniczky, Nils Gruschka, Bence Borsos, Mohamed~Amine Ferrag, Andras Kovacs, Vasileios Mavroeidis, and Norbert Tihanyi.
\newblock I know which llm wrote your code last summer: Llm generated code stylometry for authorship attribution.
\newblock arXiv preprint, 2025.
\newblock URL \url{https://arxiv.org/abs/2506.17323}.

\bibitem[Rao et~al.(2025)Rao, Mohamed, Liu, and Liu]{rao2025two}
Zhicheng Rao, Yasmine Mohamed, Sijia Liu, and Zirui Liu.
\newblock Two birds with one stone: Multitask detection and attribution of llm-generated text.
\newblock arXiv preprint, 2025.
\newblock URL \url{https://arxiv.org/abs/2508.14190}.

\bibitem[La~Cava et~al.(2025)La~Cava, Macko, Móro, Srba, and Tagarelli]{lacava2025authorship}
Lucio La~Cava, Dominik Macko, Robert Móro, Ivan Srba, and Andrea Tagarelli.
\newblock Authorship attribution in multilingual machine-generated texts.
\newblock arXiv preprint, 2025.
\newblock URL \url{https://arxiv.org/abs/2508.01656}.

\bibitem[Juola and Mikros(2016)]{juola2016cross}
Patrick Juola and George~K Mikros.
\newblock Cross-linguistic stylometric features: A preliminary investigation.
\newblock In \emph{Actes des 13èmes Journées internationales d'Analyse statistique des Données Textuelles (JADT 2016)}, pages 787--794, Nice, France, 2016.

\bibitem[Juola et~al.(2019)Juola, Mikros, and Vinsick]{juola2019comparative}
Patrick Juola, George~K Mikros, and Sean Vinsick.
\newblock A comparative assessment of the difficulty of authorship attribution in greek and in english.
\newblock \emph{Journal of the Association for Information Science and Technology}, 70\penalty0 (1):\penalty0 61--70, 2019.
\newblock \doi{10.1002/asi.24073}.

\bibitem[Dunsin et~al.(2024)Dunsin, Ghanem, Ouazzane, and Vassilev]{dunsin2024comprehensive}
Damilare Dunsin, M~C Ghanem, Karim Ouazzane, and Vassil Vassilev.
\newblock A comprehensive analysis of the role of artificial intelligence and machine learning in modern digital forensics and incident response.
\newblock \emph{Forensic Science International: Digital Investigation}, 48:\penalty0 301675, 2024.
\newblock \doi{10.1016/j.fsidi.2023.301675}.

\bibitem[Römisch et~al.(2025)Römisch, Gorovaia, Halchynska, Schmidt, and Yamshchikov]{romisch2025better}
Julian Römisch, Sofia Gorovaia, Maryna Halchynska, Georg Schmidt, and Ivan~P Yamshchikov.
\newblock Better call claude: Can llms detect changes of writing style?
\newblock arXiv preprint, 2025.
\newblock URL \url{https://arxiv.org/abs/2508.00680}.

\bibitem[Abbas(2025)]{abbas2025attribution}
M~Abbas.
\newblock Attribution quality in ai-generated content: Benchmarking style embeddings and llm judges.
\newblock arXiv preprint, 2025.
\newblock URL \url{https://arxiv.org/abs/2510.13898}.

\bibitem[Roemling et~al.(2024)Roemling, Scherrer, and Miletic]{roemling2024explainability}
David Roemling, Yves Scherrer, and Aleksandra Miletic.
\newblock Explainability of machine learning approaches in forensic linguistics: A case study in geolinguistic authorship profiling.
\newblock arXiv preprint, 2024.
\newblock URL \url{https://arxiv.org/abs/2404.18510}.

\bibitem[Alperin et~al.(2025)Alperin, Leekha, Uchendu, Nguyen, Medarametla, Capote, Aycock, and Dagli]{alperin2025masks}
Klamber Alperin, Roshni Leekha, Adaku Uchendu, Thai Nguyen, Sameera Medarametla, Christopher~L Capote, Steven Aycock, and C~Dagli.
\newblock Masks and mimicry: Strategic obfuscation and impersonation attacks on authorship verification.
\newblock arXiv preprint, 2025.
\newblock URL \url{https://arxiv.org/abs/2503.19099}.

\bibitem[Dalalah and Dalalah(2023)]{dalalah2023false}
Doraid Dalalah and Osama M~A Dalalah.
\newblock The false positives and the false negatives of generative ai detection tools in education and academic research: The case of chatgpt.
\newblock \emph{The International Journal of Management Education}, 21\penalty0 (2):\penalty0 100822, 2023.
\newblock \doi{10.1016/j.ijme.2023.100822}.

\bibitem[Rashidi et~al.(2023)Rashidi, Fennell, Albahra, Hu, and Gorbett]{rashidi2023chatgpt}
Hooman~H Rashidi, Brandon~D Fennell, Samer Albahra, Bo~Hu, and Tranassa Gorbett.
\newblock The chatgpt conundrum: Human-generated scientific manuscripts misidentified as ai creations by ai text detection tool.
\newblock \emph{Journal of Pathology Informatics}, 14:\penalty0 100342, 2023.
\newblock \doi{10.1016/j.jpi.2023.100342}.

\bibitem[Creo and Pudasaini(2024)]{creo2024silverspeak}
Andrea Creo and Shishir Pudasaini.
\newblock Silverspeak: Evading ai-generated text detectors using homoglyphs.
\newblock arXiv preprint, 2024.
\newblock URL \url{https://arxiv.org/abs/2406.11239}.

\bibitem[Zeng et~al.(2024)Zeng, Liu, Sha, Li, Yang, Liu, Gašević, and Chen]{zeng2024detecting}
Zhenghao Zeng, Sijie Liu, Lei Sha, Zhichao Li, Kailai Yang, Sijia Liu, Dragan Gašević, and Guanliang Chen.
\newblock Detecting ai-generated sentences in human–ai collaborative hybrid texts: Challenges, strategies, and insights.
\newblock arXiv preprint, 2024.
\newblock URL \url{https://arxiv.org/abs/2403.03506}.

\bibitem[Daubert(1993)]{daubert1993daubert}
Daubert v. Merrell Dow Pharmaceuticals, Inc.
\newblock 509 U.S. 579, 1993.

\bibitem[Kumho(1999)]{kumho1999kumho}
Kumho Tire Co. v. Carmichael.
\newblock 526 U.S. 137, 1999.

\bibitem[Giray(2024)]{giray2024problem}
L~Giray.
\newblock The problem with false positives: Ai detection unfairly accuses scholars of ai plagiarism.
\newblock \emph{The Serials Librarian}, 85\penalty0 (5--6), 2024.
\newblock \doi{10.1080/0361526X.2024.2433256}.

\bibitem[Zhang and Xie(2025)]{zhang2025multihierarchical}
L~Zhang and X~Xie.
\newblock Multi-hierarchical feature detection for large language model generated text.
\newblock arXiv preprint, 2025.
\newblock URL \url{https://arxiv.org/abs/2509.18862}.

\bibitem[Macko(2025)]{macko2025robustly}
Dominik Macko.
\newblock Robustly finetuned llm for binary and multiclass ai-generated text detection.
\newblock arXiv preprint, 2025.
\newblock URL \url{https://arxiv.org/abs/2506.01702}.

\bibitem[Trivedi and Sivanesan(2025)]{trivedi2025sarang}
Ansh Trivedi and S~Sivanesan.
\newblock Sarang at defactify 4.0: Detecting ai-generated text using noised data and an ensemble of deberta models.
\newblock arXiv preprint, 2025.
\newblock URL \url{https://arxiv.org/abs/2502.16857}.

\bibitem[Mikros et~al.(2023)Mikros, Koursaris, Bilianos, and Markopoulos]{mikros2023ai}
George Mikros, A~Koursaris, D~Bilianos, and G~Markopoulos.
\newblock Ai-writing detection using an ensemble of transformers and stylometric features.
\newblock In \emph{Proceedings of the Iberian Languages Evaluation Forum (IberLEF 2023)}, volume 3496, pages 1--14. CEUR Workshop Proceedings, 2023.

\bibitem[Kirchenbauer et~al.(2023)Kirchenbauer, Geiping, Wen, Shu, Saifullah, Kong, Fernando, Saha, Goldblum, and Goldstein]{kirchenbauer2023reliability}
John Kirchenbauer, Jonas Geiping, Yuxin Wen, Manli Shu, Khalid Saifullah, Kezhi Kong, Kasun Fernando, Aniruddha Saha, Micah Goldblum, and Tom Goldstein.
\newblock On the reliability of watermarks for large language models.
\newblock arXiv preprint, 2023.
\newblock URL \url{https://arxiv.org/abs/2306.04634}.

\bibitem[Verma et~al.(2025)Verma, Phan, and Trivedi]{verma2025watermarking}
Akshat Verma, Nhat Phan, and S~Trivedi.
\newblock Watermarking degrades alignment in language models: Analysis and mitigation.
\newblock arXiv preprint, 2025.
\newblock URL \url{https://arxiv.org/abs/2506.04462}.

\bibitem[Xu et~al.(2025)Xu, Liu, Hu, Wen, and Xiong]{xu2025mark}
Yige Xu, Aling Liu, Xiting Hu, L~Wen, and Hui Xiong.
\newblock Mark your llm: Detecting the misuse of open-source large language models via watermarking.
\newblock arXiv preprint, 2025.
\newblock URL \url{https://arxiv.org/abs/2503.04636}.

\bibitem[Cao(2025)]{cao2025watermarking}
L~Cao.
\newblock Watermarking for ai content detection: A review on text, visual, and audio modalities.
\newblock arXiv preprint, 2025.
\newblock URL \url{https://arxiv.org/abs/2504.03765}.

\bibitem[Zhang et~al.(2025)Zhang, Liu, Liu, Gao, Li, Gu, and Hu]{zhang2025cohemark}
J~Zhang, S~Liu, A~Liu, Y~Gao, J~Li, X~Gu, and X~Hu.
\newblock Cohemark: A novel sentence-level watermark for enhanced text quality.
\newblock arXiv preprint, 2025.
\newblock URL \url{https://arxiv.org/abs/2504.17309}.

\bibitem[Cheng et~al.(2024)Cheng, Zhou, Jiang, Wang, and Li]{cheng2024beyond}
Z~Cheng, L~Zhou, F~Jiang, B~Wang, and H~Li.
\newblock Beyond binary: Towards fine-grained llm-generated text detection via role recognition and involvement measurement.
\newblock arXiv preprint, 2024.
\newblock URL \url{https://arxiv.org/abs/2410.14259}.

\bibitem[Qazi et~al.(2024)Qazi, Shiao, and Papalexakis]{qazi2024gpt}
Z~Qazi, W~Shiao, and E~E Papalexakis.
\newblock Gpt-generated text detection: Benchmark dataset and tensor-based detection method.
\newblock arXiv preprint, 2024.
\newblock URL \url{https://arxiv.org/abs/2403.07321}.

\end{thebibliography}
\end{document}